%% file: main.tex
\pdfoutput=1

\documentclass[11pt]{article}

\usepackage{acl}

\usepackage{times}
\usepackage{latexsym}

\usepackage[T1]{fontenc}

\usepackage[utf8]{inputenc}

\usepackage{microtype}

\usepackage{url}            
\usepackage{booktabs}       
\usepackage{amsmath}
\usepackage{amsfonts}       
\usepackage{nicefrac}       
\usepackage{natbib}
\usepackage{array}
\usepackage{makecell}
\usepackage{multirow}
\usepackage{amsfonts}
\usepackage{graphicx}
\usepackage{wrapfig} 
\usepackage{float}
\usepackage{caption} 
\usepackage{mathtools}
\usepackage[autostyle]{csquotes}
\usepackage{xcolor}
\usepackage{caption}
\usepackage{subcaption}
\usepackage{multirow}
\usepackage{enumitem}
\usepackage{xspace}

\newcolumntype{M}[1]{>{\centering\arraybackslash}m{#1}}
\newcolumntype{N}{@{}m{0pt}@{}}
\newcolumntype{P}[1]{>{\centering\arraybackslash}p{#1}}

\colorlet{myGreen}{green!40!gray}
\newcommand{\NoGap}{\vspace{-5mm}}

\newcommand{\TablePad}{\vspace{-2mm}}

\newcommand{\loo}{\textsc{l1o}}

\newcommand{\adasumnoperc}{\textsc{AdaSum}}
\newcommand{\adaqsumnoperc}{\textsc{AdaQSum}} 

\newcommand{\adasum}[1]{\textsc{AdaSum (#1\%)}} 
\newcommand{\adasumloo}[1]{\textsc{AdaSum (#1\%) + l1o}} 

\newcommand{\adaqsum}[1]{\textsc{AdaQSum (#1\%)}} 
\newcommand{\adaqsumloo}[1]{\textsc{AdaQSum (#1\%) + l1o}} 

\makeatletter
\newcommand{\thickhline}{%
    \noalign {\ifnum 0=`}\fi \hrule height 1pt
    \futurelet \reserved@a \@xhline
}
\makeatother

\title{Efficient Few-Shot Fine-Tuning for Opinion Summarization}

\author{Arthur Bražinskas$^1$ \, Ramesh Nallapati$^2$ \, Mohit Bansal$^{2,3}$ \, Markus Dreyer$^2$
\\
$^1$ILCC University of Edinburgh \,
$^2$Amazon \\
$^3$UNC Chapel Hill\\
\texttt{abrazinskas@ed.ac.uk}, \texttt{\{rnallapa, mobansal, mddreyer\}@amazon.com}
}

\date{}

\begin{document}

\maketitle

\input{content/sections/abstract}

\input{content/sections/introduction}

\input{content/sections/approach}

\input{content/sections/exp_setup}

\input{content/sections/results}

\input{content/sections/analysis}

\input{content/sections/related_work}

\input{content/sections/conclusion}

\input{content/sections/acknowledgment}

\input{content/sections/ethics}

\input{content/sections/limitations}

\bibliographystyle{acl_natbib} 
\bibliography{anthology,custom} 

\input{content/appendix/main}

\end{document}

%% file: content/sections/abstract.tex
\begin{abstract}
Abstractive summarization models are typically pre-trained on large amounts of generic texts, then fine-tuned on tens or hundreds of thousands of annotated samples. However, in opinion summarization, large annotated datasets of reviews paired with reference summaries are not available and would be expensive to create. This calls for fine-tuning methods robust to overfitting on small datasets. In addition, generically pre-trained models are often not accustomed to the specifics of customer reviews and, after fine-tuning, yield summaries with disfluencies and semantic mistakes. To address these problems, we utilize an efficient few-shot method based on adapters which, as we show, can easily store in-domain knowledge. Instead of fine-tuning the entire model, we add adapters and pre-train them in a task-specific way on a large corpus of unannotated customer reviews, using held-out reviews as pseudo summaries. Then, fine-tune the adapters on the small available human-annotated dataset. We show that this self-supervised adapter pre-training improves summary quality over standard fine-tuning by 2.0 and 1.3 ROUGE-L points on the Amazon and Yelp datasets, respectively. Finally, for summary personalization, we condition on aspect keyword queries, automatically created from generic datasets. In the same vein, we pre-train the adapters in a query-based manner on customer reviews and then fine-tune them on annotated datasets. This results in better-organized summary content reflected in improved coherence and fewer redundancies.
\end{abstract}

%% file: content/sections/introduction.tex
\section{Introduction}
\label{sec:introduction}

\begin{figure*}[t!]
   \vspace{-35pt}
    \centering
    \includegraphics[width=0.85\textwidth]{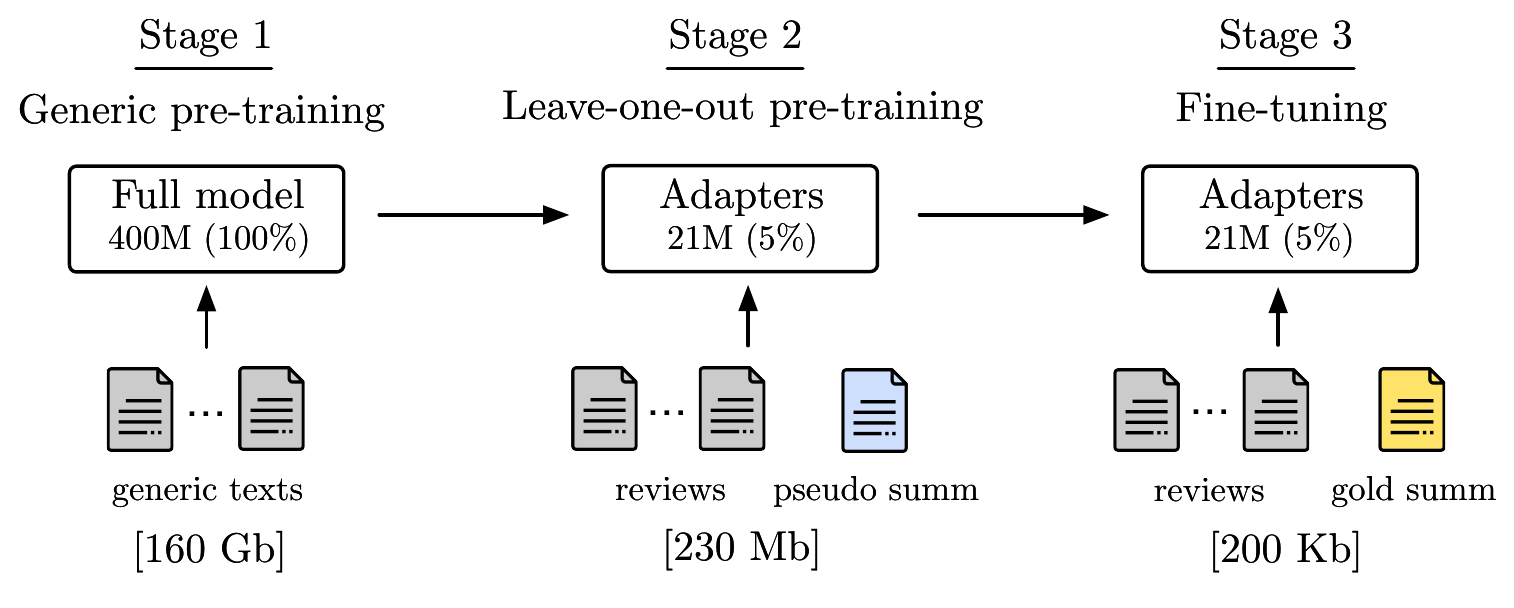}
    \vspace*{-4mm}
    \caption{Illustration of the proposed approach. In Stage 1, all parameters of a large language model are pre-trained on generic texts (we use \textsc{BART}). In Stage 2, we pre-train adapters (5\% of the full model's parameters) on customer reviews using held-out reviews as summaries. In Stage 3, we fine-tune the adapters on a handful of reviews-summary pairs. 
    }
    \label{fig:workflow}
    \TablePad{}
\end{figure*}

Online reviews play an important role in purchasing decisions we make. They inform us about customer experiences -- what aspects users like and dislike, and ultimately, whether a product or service is worth buying.  Although significant progress has been made in supervised summarization in non-subjective single-document context, such as news articles~\citep{rush2015neural, nallapati2016abstractive, see2017get, bravzinskas2021transductive}, modern deep learning methods rely on large amounts of annotated data, but these are not readily available in the
review or opinion summarization domain and are expensive to produce. A key 
obstacle making such annotation expensive is that annotators need to consider multiple texts when writing a summary, which can be tens or even hundreds for realistic settings. Hence, most available datasets have human-written summaries for less than 100 products. 
\vspace{3.0mm}
\newline
\indent
The lack of sufficiently large annotated datasets led to a variety of unsupervised abstractive models (e.g., \textsc{Copycat}~\citep{brazinskas2020-unsupervised}, \textsc{MeanSum}~\citep{chu2019meansum}; \textsc{DenoiseSum}~\citep{amplayo2020unsupervised}) that are trained on large collections of unannotated customer reviews. However, as the models are never exposed to actual summaries, they cannot capture their expected characteristics. This results in generated summaries mimicking the informal style of customer reviews and containing hallucinations and unimportant details.
These limitations were addressed by few-shot methods learning from a handful of human-written summaries~\citep{brazinskas2020few, Oved2021PASSPS}. The first proposed few-shot learning model, \textsc{FewSum}~\citep{brazinskas2020few}, relies on various static features capturing differences between a customer review and a summary. While that model is more robust to overfitting, such features require manual, domain-specific engineering and can be sub-optimal for capturing correspondences between texts on the semantic level. We propose a simpler approach -- \adasumnoperc{} -- which is based on adapters~\citep{houlsby2019parameter, bapna2019simple}. As we explain next, the adapters are pre-trained on customer reviews in a task-specific manner, and subsequently fine-tuned on gold summaries. 

\paragraph{Adapters.}
We utilize a pre-trained model with powerful language understanding and generation abilities in combination with a parameter-efficient fine-tuning method -- adapters. As was shown in recent studies, this method is also robust to overfitting in low-resource settings~\citep{he-etal-2021-effectiveness}. In this way, a large pre-trained model, BART~\citep{lewis2019bart}, in our case, remains frozen, and only small modules (0.6\% - 5\% of the model parameters) are optimized. This effectively retains acquired knowledge in the pre-trained language model (PLM) without specialized training objectives as in \textsc{RecAdam}~\citep{chen-etal-2020-recall}. However, available annotated data is not sufficient for learning in-domain specifics and results in summaries with subtle semantic mistakes. As explained next, we reduce these semantic mistakes by pre-training adapters on customer reviews.

\paragraph{Self-supervised Pre-training.} 

Language models are pre-trained with generic objectives (e.g., single-document denoising) and rarely on in-domain data, such as customer reviews. Consequently, this makes them less attuned to in-domain specifics as these are hard to learn from a handful of summaries. This often results in subtle semantic mistakes. For instance, in Table~\ref{table:front_summary}, \textsc{PASS}~\citep{Oved2021PASSPS} incorrectly concludes that \textit{thin material} implies \textit{poor quality}. To address this issue, we learn in-domain specifics from customer reviews. Concretely, we employ a self-supervised pre-training method: For any given product without a human-written summary, we predict one of the given reviews by conditioning on $N$ other reviews with the highest lexical overlap in the \textit{leave-one-out} fashion~\citep{brazinskas2020-unsupervised}. As the standard training of PLMs is storage and memory inefficient~\citep{mahabadi2021compacter}, we pre-train adapters only; see Stage 2 in Fig.~\ref{fig:workflow}. Afterwards, we fine-tune them on a small number of annotated reviews-summary pairs (< 100 pairs), see Stage 3 in Fig.~\ref{fig:workflow}. All in all, our method combines the general text generation and understanding abilities of the PLM with in-domain knowledge directly related to the end task. 

\paragraph{Content Planning.}
Well-organized content in summaries is easier to follow and thus improves user experience. However, the lack of annotated data makes it challenging to learn a desired content structure. For example, in Table~\ref{table:front_summary}, \textsc{FewSum}'s summary does not end after a concluding phrase `\textit{Other than that, it’s a great top.}' While the state-of-the-art model (\textsc{PASS}) addresses this issue by ranking multiple generated candidates with a specialized coherence model, we propose a simpler solution -- \adaqsumnoperc{} -- that capitalizes on \textit{text planning}~\citep{hua-wang-2019-sentence, step-by-step}. Specifically, we allow the model to \textit{plan ahead} by providing an intermediate summary representation in the form of a query consisting of aspect keywords. As we show, this results in more coherent text patterns with fewer redundancies. Moreover, it can be useful for personalized summaries, better reflecting user interests.

\input{content/examples/front_summ}

\paragraph{Result Highlights.}
We evaluate the proposed models in terms of automatic metrics and human efforts. We find that pre-training and fine-tuning of adapters leads to more than 2.0 and 1.3 ROUGE-L points improvement over fine-tuning the entire model on Amazon and Yelp datasets, respectively. We also find that our pre-trained and fine-tuned query-based model improves ROUGE-L scores by more than 2.7 and 0.9 ROUGE-L points over \textsc{PASS}, on Amazon and Yelp datasets, respectively, and is more preferred by humans. We further demonstrate that the query-based model (\adaqsumnoperc{}) substantially improves coherence and reduces redundancies in generated summaries.

\vspace{2mm}\noindent In summary, our contributions are as follows:\vspace{-3mm}

\begin{itemize}[leftmargin=*]
  \setlength\itemsep{-0.2em}
    \item We propose a self-supervised pre-training method to learn in-domain knowledge by adapters that alleviate catastrophic forgetting;
    \item We propose, to the best of our knowledge, the first aspect-based abstractive opinion summarizer learned from a few annotated samples;
    \item We substantially increase summary coherence using the query-based approach; 
    \item We show that self-supervised pre-training significantly improves performance on the query-based task;
    \item We demonstrate state-of-the-art results on two primary benchmarks in automatic and human evaluation.\footnote{Our code and associated artifacts is publicly available at \url{https://github.com/amazon-research/adasum}.}
\end{itemize}

%% file: content/examples/front_summ.tex
\begin{table}[t!]
    \centering
 	\footnotesize 
    \begin{tabular}{ >{\centering\arraybackslash} m{1.4cm} m{5.6cm}} 
 \thickhline 
 \textsc{FewSum} & \vspace{0.5em} This tank top is well made, fits well, and is comfortable to wear. The only thing is that it runs a little small, so order a size up from what you normally wear. Other than that, it's a great top. It's well made and it looks like it will last a long time. Love it!  \vspace{0.5em} \\ \hline
     \textsc{PASS} & \vspace{0.5em}  This is a basic tank. The photo shows it going well past the models hips. However, the material used to make it this long is thin and therefore not good quality. It is also thinner than other tanks on the market but is still comfortable to wear.  \vspace{0.5em} \\  \hline
    \textsc{AdaQSum} & \vspace{0.5em} This is a basic tank top that \textcolor{orange}{fits} well and is comfortable to \textcolor{blue}{wear}. The \textcolor{cyan}{color} is great and the length is long enough to \textcolor{blue}{wear} with leggings. The \textcolor{red}{quality} of the product is good. \vspace{0.5em} \\
    \thickhline
    \textsc{Reviews} & \vspace{0.5em} ... This is a basic tank ... || ... this tank \textcolor{orange}{fits} like a normal tank top, not any longer ... I could \textcolor{blue}{wear} it with leggings ... || ... It is THIN and runs SMALL ... It \textcolor{orange}{fits} tight and is NOT long like in the picture ... || The tank \textcolor{orange}{fit} very well and was comfortbale to \textcolor{blue}{wear}. I've bought much higher \textcolor{red}{quality} tanks ... || ... it is listed as a 'long' tank top and the photo even shows it going well past the models hips, however I'm short and the tank top is just a normal length. || ... They were a lot thinner than I like ... || Every women should own one in every \textcolor{cyan}{color}. Just feels \textcolor{red}{quality} I don't know how else to explain it ... || ... They are long enough that the \textcolor{cyan}{color} peeks out from under my tops. Looks cute.
    \vspace{0.5em} \\ \thickhline
    \end{tabular}
    \caption{Generated summaries for an Amazon product by baseline models (\textsc{FewSum} and \textsc{Pass}) and our approach (\textsc{AdaQSum}). Colored words indicate aspect keywords that were part of the query. The special marker `||' separates truncated reviews.
    }
    \label{table:front_summary}
     \NoGap{}
\end{table}

%% file: content/sections/approach.tex
\section{Approach}
\label{sec:approach}

\subsection{Opinion Summarization Tasks}
\label{sec:generic_and_query_based_summ}
\begin{figure}[t!]
    \vspace{-5.5mm}
    \centering
    \includegraphics[width=0.49\textwidth]{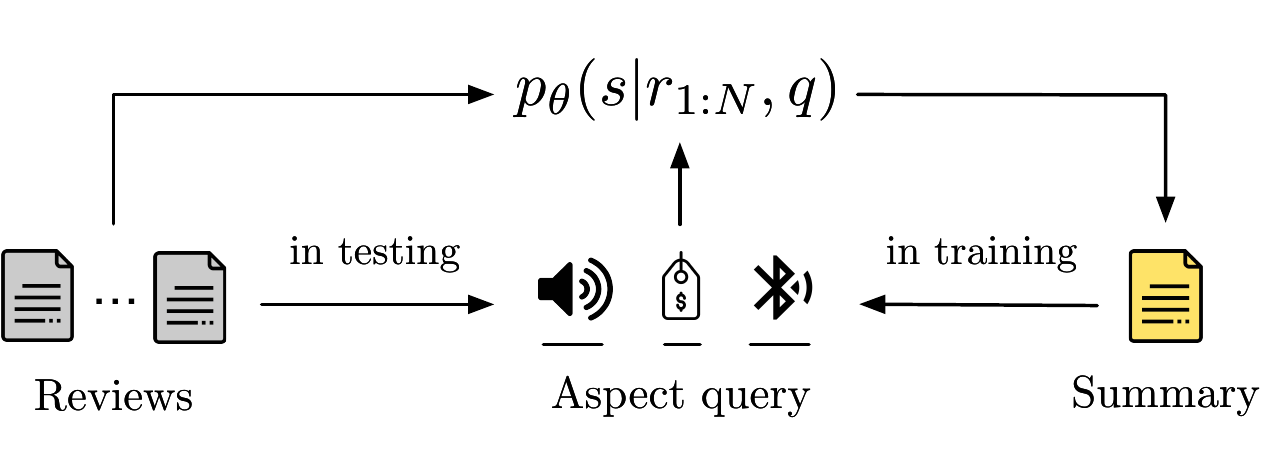}
    \caption{Illustration of the query-based summarizer that inputs reviews and a text query consisting of aspects, such as `volume,' `price,' and `bluetooth.' The query is automatically created from gold summaries in training and reviews in test time.}
    \label{fig:query_based}
    \vspace{-0.8em}
\end{figure}

In this work, we consider two tasks of customer review summarization. The first one is \textit{generic summarization}~\citep{chu2019meansum, brazinskas2020few}, where the aim is to produce a summary that covers overall opinions in input reviews. Formally, given $N$ input reviews $r_{1:N}$, the task is to predict word-by-word the summary $s$:
\vspace{-0.5em}
\begin{equation*}
    \mathcal{L}(s, r_{1:N}; \theta) = \sum_{t=1}^T \log p_{\theta}(s^t|s^{1:{t-1}}, r_{1:N}).
    \label{eq:ll}
\end{equation*}
In the second task, \textit{query-based summarization}, we assume that the user provides a query $q$ consisting of aspect keywords, such as `bluetooth,' `resolution,' and `battery life.' In turn, a summarizer should generate a summary reflecting customer opinions in $r_{1:N}$ about these aspects. Formally, given a pair of input reviews and query ($r_{1:N}$, $q$), the task is to predict word-by-word the summary $s$:
\vspace{-0.5em}
\begin{equation*}
    \mathcal{L}(s, r_{1:N}, q; \theta) = \sum_{t=1}^T \log p_{\theta}(s^t|s^{1:{t-1}}, r_{1:N}, q).
    \label{eq:ll_q}
\end{equation*}
Unfortunately, abstractive opinion datasets with annotated aspect queries are unavailable in the domain. To mitigate this problem, we follow \citet{ni-etal-2019-justifying} and create queries by extracting fine-grained aspect keywords from available generic summaries. Specifically, we utilize the model proposed by \citet{zhang2014explicit} to build a fine-grained aspect lexicon from review datasets. Further, we use simple rules to determine which aspects appear in summaries; see an annotated summary in Table~\ref{table:summary_annotated}. At test time, we follow the intuition that a summary should reflect common opinions and create a query from $K$ most frequent aspect keywords in input reviews. The workflow is illustrated in Fig.~\ref{fig:query_based}.

\subsection{Model}
\label{sec:model}

Our model is based on the Transformer~\citep{vaswani2017attention} encoder-decoder architecture initialized with BART~\citep{lewis2019bart}.  We adopt the same encoder as in \citep{Raffel2020ExploringTL, Oved2021PASSPS} where reviews are concatenated before encoding.\footnote{We experimented with the independent review encoding as in  \citet{brazinskas2020-unsupervised}. However, the results were slightly worse.} This allows us to capture product-level features and leverage commonalities across reviews during encoding. For query-based summarization, we concatenate a query and reviews while indicating boundaries with special markers. In this way, the encoder can contextualize aspect keywords and focus on salient review fragments reflecting these aspects. 

\subsection{Adapters}
\label{sec:adapters}
In training, a large pre-trained model remains frozen and only tiny neural networks called \textit{adapters}~\citep{houlsby2019parameter} are optimized. These modules are injected into the transformer layers (both encoder and decoder). Formally, given the input hidden vector $h$, the output vector $\hat{h}$ is calculated as shown below:
\vspace{-0.3em}
\begin{equation*}
    \hat h = f_2(\tanh f_1(h)) + h
    \label{eq:adapters} .
\end{equation*}

The functions $f_1(\cdot)$ and $f_2(\cdot)$ are the down- and up- projection layers. At each transformer layer, two adapters are inserted right after the self-attention and the feed-forward layers, respectively. These modules consist of substantially fewer parameters than the language model, usually around 3\% - 5\%. Recent studies have shown that adapters are less prone to overfitting~\citep{he-etal-2021-effectiveness} and are more memory-efficient in training~\citep{mahabadi2021compacter}. Finally, as the pre-trained model remains frozen, it retains all the prior knowledge for text understanding and generation. This effectively alleviates catastrophic forgetting~\citep{goodfellow2013empirical, kemker2017measuring} without modifying the training objective as in \textsc{RecAdam}~\citep{chen-etal-2020-recall, yu2021AdaptSum}. We refer to our approaches as \adasumnoperc{} and \adaqsumnoperc{} for generic and query-based summarization, respectively.

\input{content/examples/annotated_summary}

\subsection{Self-supervised Pre-training}
\label{sec:self_sup_pretraining}
Language models, initially pre-trained on generic text corpora, are often not accustomed to in-domain specifics. Unsurprisingly perhaps, a wide range of product-related specifics cannot be learned from a handful of annotated summaries during fine-tuning. Consequently, this can result in subtle semantic mistakes in generated summaries. We will discuss this problem and provide examples in Sec.~\ref{sec:semantic_mistakes}. Furthermore, query-based summarization is even more challenging for learning than generic summarization. To be useful in practice, the summarizer should rely on the provided query after fine-tuning. However, a handful of annotated samples might be insufficient to learn this dynamic. We will analyze this problem in Sec.~\ref{sec:query_based_summarization_results}. To alleviate these two problems, we leverage unannotated customer reviews to construct synthetic datasets for pre-training. 

\paragraph{Synthetic In-Domain Pre-Training Dataset.}
\label{sec:synthetic_dataset}

From a group of product reviews, we randomly sample one review as a \textit{pseudo summary} $s$ and select $N$ reviews as input ($r_{1:N}$).\footnote{We also experimented with selecting pseudo summaries without personal pronouns -- written in the formal style~\citep{brazinskas2020few}. However, we did not observe significant improvements.} We select $N$ input reviews covering the content of the summary $s$ -- that have the highest ROUGE-1 F scores. Following the naming convention~\citep{brazinskas2020-unsupervised}, we refer to this as \textit{leave-one-out pre-training} (\loo{}). To closely resemble query-based summarization, we create aspect queries from pseudo summaries. Specifically, we leverage the aspect lexicon by matching summary keywords; in the same way as was explained in Sec.~\ref{sec:generic_and_query_based_summ}. In practice, we expect queries to have at least one aspect keyword. Therefore, we remove all pre-training pairs where the pseudo summary has no aspect keywords.

%% file: content/examples/annotated_summary.tex
\begin{table}[t]
    \centering
    \footnotesize
    \begin{tabular}{ >{\arraybackslash} m{7cm}} \hline 
         \vspace{0.5em}
         The \underline{\textit{cover}} offers durable \underline{\textit{protection}} for the MacBook, the retractable \underline{\textit{tilt}} stands offer \underline{\textit{protection}} for the \underline{\textit{wrists}}.\\ \vspace{0.25em}
         The \underline{\textit{keyboard cover}} can take some effort to \underline{\textit{fit}} properly, and \underline{\textit{adjustment}} to its feel may take time. \\\vspace{0.25em}
         However, free and fast \underline{\textit{shipping}} make up for this one potential issue. \vspace{0.5em} \\ \hline 
    \end{tabular}
    \caption{Automatically annotated Amazon summary with fine-grained aspect keywords (\underline{\textit{underlined italic}}).}
    \label{table:summary_annotated}
    \TablePad{}
\end{table}

%% file: content/sections/exp_setup.tex
\section{Experimental Setup}
\label{sec:experimental_setup}

\subsection{Data}
\label{sec:data}

To create synthetic datasets, we used customer reviews from Amazon \citep{he2016ups} and Yelp.\footnote{\url{https://www.yelp.com/dataset}}  Following \citet{brazinskas2020few}, we selected 4 categories: \textit{Electronics}; \textit{Clothing, Shoes and Jewelry}; \textit{Home and Kitchen}; \textit{Health and Personal Care}. We pre-processed the datasets by removing all reviews that are shorter than 20 words and longer than 120 words and evened the number of pairs in both datasets. Further, we used Amazon and Yelp gold summaries from \citep{brazinskas2020few} where each product/business has 3 references and is paired with 8 reviews. Gold and synthetic dataset statistics\footnote{For the query-based setup, we removed all instances where targets had no aspects.} are presented in Table~\ref{table:data_statistics}. 

\begin{table}[t]
    \centering
    \small
    \begin{tabular}{ c | c | c | c | c }
        \multicolumn{1}{c}{} & \multicolumn{2}{c}{\textbf{Amazon}} & \multicolumn{2}{c}{\textbf{Yelp}} \\ \thickhline
    Split & Gold & Synthetic & Gold & Synthetic \\ \thickhline
    Train & 84 & 70,144\;/\;59,963 & 90 & 70,144\;/\;68,499 \\
    Valid & 36 & 7,900\;/\;6,810 & 90 & 7,900\;/\;7,724\\ 
    Test & 60 & - & 120 & - \\ \hline
    \end{tabular}
    \caption{Source-target pair numbers for Amazon and Yelp, both gold and synthetic. Each pair has 8 source reviews. Generic and query-based pair statistics are separated by '/`.}
    \label{table:data_statistics}
    \TablePad{}
\end{table}

\subsection{Baselines}
\label{sec:baselines}
\textsc{LexRank}~\citep{erkan2004lexrank} is an unsupervised extractive graph-based model that selects sentences based on graph centrality. Sentences represent nodes in a graph whose edges are weighted with tf-idf. 

\textsc{MeanSum}~\citep{chu2019meansum} is an unsupervised abstractive summarization model which treats a summary as a structured latent state of an auto-encoder trained to reconstruct reviews of a product. 

\textsc{Copycat}~\citep{brazinskas2020-unsupervised} is the state-of-the-art unsupervised abstractive summarizer with hierarchical continuous latent representations to model products and individual reviews.

\textsc{FewSum}~\citep{brazinskas2020few} is a few-shot framework where lexical features are used to differentiate between customer reviews and summaries. In the fine-tuning phase, features leading to generation of summaries are searched. 

\textsc{PASS}~\citep{Oved2021PASSPS} is based on a pre-trained T5 model~\citep{Raffel2020ExploringTL} that is further fine-tuned on gold summaries. At inference, the model's input is perturbed to generate multiple candidates. These candidates are further ranked by a separate model based on coherence and fluency to select the best one. 

We fine-tuned the full BART model (\textsc{Full}) for a fair comparison, with and without the leave-one-out pre-training. We also employed a number of simple summarization baselines. First, the \textsc{clustroid} review was computed for each group of reviews as follows. We took each review from a group and computed ROUGE-L with respect to all other reviews. The review with the highest ROUGE score was selected as the clustroid review. Second, we sampled a \textsc{random} review from each group to be used as the summary. And lastly, we constructed the summary by selecting the \textit{leading sentences} (\textsc{Lead}) from each review of a group.

\subsection{Experimental Details}
\label{sec:exper_details}

We used a standard Transformer encoder-decoder~\citep{vaswani2017attention}, pre-initialized with \textsc{BART} large~\citep{lewis2019bart}, consisting of 400M parameters. We used two adapter sizes -- 0.6\% and 5\% of the full model's parameters. All input reviews were concatenated, following \citet{Raffel2020ExploringTL, Oved2021PASSPS}. For parameter optimization, we used \textsc{Adam}~\citep{kingma2014adam}, and summary generation was performed via the beam search of size 5 and with 3-gram blocking~\citep{paulus2017deep}. We used ROUGE-L as the stopping criterion on the end task, and perplexity (PPL) for pre-training. The learning rate for most experiments was set to 5e-5. Aspect lexicons for query-based summarization contained 2809 and 4013 fine-grained aspects for Amazon and Yelp, respectively. In pre-training and fine-tuning, we shuffled aspects to break temporal dependencies. For fine-tuning on Yelp, we also found it useful to exclude summary aspect keywords that do not appear in input reviews. This approximately matched the number of aspect keywords for Amazon and Yelp. At test time, we selected up to 6 and 5 most frequent aspects for Amazon and Yelp, respectively. All computations were performed on an 8-GPU p3.8-xlarge Amazon instance.

%% file: content/sections/results.tex
\section{Results}
\label{sec:results}

\subsection{Automatic Evaluation}
\label{sec:automatic_evaluation}

\begin{table*}[h!]
    \centering
    \small
    \begin{tabular}{ l | c | c c c c | c c c c}
    \multicolumn{1}{c}{} & \multicolumn{1}{c}{} & \multicolumn{4}{c}{\textbf{Amazon}} & \multicolumn{4}{c}{\textbf{Yelp}} \\ \thickhline
    & Params$\downarrow$ & PPL$\downarrow$ & R1$\uparrow$ & R2$\uparrow$ & RL$\uparrow$ & PPL$\downarrow$ & R1$\uparrow$ & R2$\uparrow$ & RL$\uparrow$ \\
    \thickhline
    \multicolumn{7}{c}{} \\ \thickhline 
    \textsc{Clustroid} & - & - & 27.16 & 3.61 & 16.77 & - & 28.90 & 4.90 & 18.00 \\
    \textsc{Lead} & - & - & 27.00 & 4.92 & 14.95 & - & 26.20 & 4.57 & 14.32 \\
    \textsc{Random} & - & - & 25.00 & 3.82 & 15.72 & - & 21.48 & 2.59 & 13.87 \\ \thickhline 
    \multicolumn{9}{c}{\textit{Unsupervised}} \\ \thickhline 
    \textsc{LexRank}~\citep{erkan2004lexrank} & - & - & 27.72 & 5.06 & 17.04 & - & 26.96 & 4.93 & 16.13 \\
    \textsc{MeanSum}~\citep{chu2019meansum} & 25M & - & 26.63 & 4.89 & 17.11 & - & 27.50 & 3.54 & 16.09 \\
    \textsc{Copycat}~\citep{brazinskas2020-unsupervised} & 25M & - & 27.85 & 4.77 & 18.86 & - & 28.12 & 5.89 & 18.32 \\ 
    
    \thickhline 
    \multicolumn{9}{c}{\textit{Few-shot}} \\ \thickhline 
    \textsc{FewSum}~\citep{brazinskas2020few} & 25M & - & 33.56 & 7.16 & 21.49 & - & 37.29 & 9.92 & 22.76 \\
    \textsc{PASS}~\citep{Oved2021PASSPS} & 440M & - & 37.43 & 8.02 & 23.34 & - & 36.91 & 8.12 & 23.09 \\ 
    \textsc{Full (100\%)} & 400M & 17.87 & 37.22 & 9.17 & 23.51 & 12.87 & 37.40 & 10.27 & 23.76 \\
    \textsc{Full (100\%) + l1o} & 400M & 16.90 & 37.67 & 10.28 & 24.32 & 12.40 & 36.79 & 11.07 & 25.03 \\
    \adasum{0.6} & 2.6M & 13.45 & 38.49 & 9.84 & 24.37 & 11.94 & 37.55 & 10.11 & 24.08 \\
    \adasumloo{0.6} & 2.6M & 12.06 & 38.94 & 10.63 & 24.95 & 11.23 & 37.78 & 11.31 & 24.04 \\
    \adasum{5} & 21.3M & 16.30 & 38.15 & 9.18 & 23.17 & 12.50 & 38.12 & 10.89 & 24.11 \\
    \adasumloo{5} & 21.3M & \textbf{12.03} & \textbf{39.78} & \textbf{10.80} & \textbf{25.55} & \textbf{11.11} & \textbf{38.82} & \textbf{11.75} & \textbf{25.14} \\
    \thickhline
    \end{tabular}
    \caption{
    Test set ROUGE F1 scores on gold Amazon and Yelp datasets for generic review summarization. \textsc{l1o} stands for leave-one-out pre-training. We also provide the total number of trainable parameters.}
    \label{table:auto_res_generic_summ}
    \TablePad{}
\end{table*}

Table~\ref{table:auto_res_generic_summ} shows results on the Amazon and Yelp test sets for generic summarization. It shows ROUGE F1 scores~\citep{lin2004rouge} as a standard measure of informativeness\footnote{For consistency with previous works, we used the same Python package (\url{https://github.com/google-research/google-research/tree/master/rouge})} and perplexity (PPL) as a measure of confusion. 

First of all, the results indicate the superiority of adapters over full fine-tuning and state-of-the-art few-shot models on both datasets. As was observed in~\citet{he-etal-2021-effectiveness}, adapters are less prone to overfitting, which is especially beneficial in few-shot settings. Second, we observe a significant improvement in ROUGE scores when pre-trained models are further trained using \loo{}. This signifies the importance of learning in-domain specifics before fine-tuning. We also observe that adapters are more effective on the Amazon dataset, which is more challenging as indicated by higher perplexity (PPL).\footnote{Training sets are of similar sizes, i.e., 84 and 90 summaries on Amazon and Yelp, respectively} We hypothesize that the pre-trained language model (\textsc{BART}) is more accustomed to restaurant- than product-related texts. Moreover, larger adapters (5\%) tend to overfit on the small number of annotated instances, and \loo{} pre-training helps substantially, as indicated both by ROUGE scores and PPL. We provide example generated summaries in the Appendix.

\subsection{Human Evaluation}
\label{sec:human_eval}

\paragraph{Coherence Improvement.}

\begin{table*}[t!]
\centering
\small
\begin{tabular}{l l c c c c c} \thickhline
        &  & R1 & R2 & RL & unique 1-gram (\%) & unique 2-gram (\%)  \\ \thickhline
      \multirow{2}{*}{Amazon} & \adasumloo{5} & \textbf{39.78} & \textbf{10.80} & 25.55 & 67.72 & 80.83 \\ 
        & \adaqsumloo{5} & 38.53 & 10.52 & \textbf{26.06} & \textbf{69.38} & \textbf{82.57} \\ \hline
        
        \multirow{2}{*}{Yelp} & \adasumloo{5} & \textbf{38.82} & \textbf{11.75} & \textbf{25.14} & 62.26 & 76.55 \\
        & \adaqsumloo{5} & 36.79 & 10.06 & 23.99 & \textbf{65.74} & \textbf{79.88} \\ 
      \thickhline
\end{tabular}
    \caption{Comparison of the query-based and generic summarizers on test sets. Unique n-grams were computed in generated summaries.}
    \label{table:query_based_auto_and_human}
    \TablePad{}
\end{table*}

As was observed in \citep{Oved2021PASSPS}, opinion summarizers sometimes generate incoherent summaries. We hypothesized that a query should allow the model to plan ahead of time and thus generate more coherent and less redundant texts. To test the hypothesis, we compared 5\% adapter-based models with and without the query; both were pre-trained via \loo{}. 
We performed human evaluation in terms of \textit{coherence} and \textit{non-redundancy} via Best-Worst Scaling (BWS)~\citep{louviere1991best, louviere2015best}. BWS has been shown to produce more reliable results than ranking scales~\citep{kiritchenko-mohammad-2016-capturing}. 

For each Amazon test set entry and criterion, we asked three independent workers on Amazon Mechanical Turk (AMT) to select the best and worst summary. For each criterion, a system’s score is computed as the percentage of times it was selected as best, minus the percentage of times it was selected as worst~\citep{orme2009maxdiff}. The scores range from -100 (unanimously worst) to +100 (unanimously best). For more details, please refer to Appendix~\ref{app:best_worst}. 

First, the results indicate that the summaries generated by \adaqsumnoperc{} are substantially more preferred to \adasumnoperc{} in terms of coherence and non-redundancy. Namely, +13.73 vs -30.91 and -1.96 vs -25.93 for coherence and non-redundancy, respectively. We also computed the percentage of unique n-grams in each generated summary for both datasets, as shown in Table~\ref{table:query_based_auto_and_human}. The results support that query-based summaries are less redundant. However, similar to findings in \citet{Oved2021PASSPS}, we observe that more coherent summaries tend to get lower ROUGE scores. Nevertheless, our model outperforms \textsc{PASS} by a margin on both datasets -- by 2.72 and 0.9 ROUGE-L points on Amazon and Yelp, respectively.

\paragraph{Comparison to Baselines.}

To understand better how our query-based model compares to other models, we performed an additional human evaluation experiment. We used the following criteria: \textit{coherence}, \textit{non-redundancy}, and \textit{fluency}. As previously, we used the Best-Worst scaling on the Amazon test set. We assigned three AMT workers to each tuple containing summaries from \textsc{PASS}, \adaqsumloo{5}, \textsc{LexRank}, and human annotators (\textsc{Gold}). 

\begin{table}[t]
\centering
\small
\begin{tabular}{l c c c c c} \thickhline
       & Fluency & Coher. & Non-Red.  \\ \thickhline
       \textsc{PASS} & -21.74 & \textbf{+33.33} & 0.00 \\ 
       \textsc{LexRank} & -45.95 & -52.38 & -58.97\\ 
       \adaqsumloo{5} & \textbf{+26.67} & +25.00 & \textbf{+26.67} \\ 
       \thickhline
       \textsc{Gold} & +46.67 & +27.78 & +55.56 \\ \thickhline
\end{tabular}
    \caption{Human evaluation results in terms of the Best-Worst scaling on the Amazon test set.}
    \label{table:ama_human_eval}
    \TablePad{}
\end{table}

The results in Table~\ref{table:ama_human_eval} suggest that summaries produced by our model are more fluent and non-redundant than the ones produced by \textsc{PASS}. In general, \textsc{PASS} produces more diverse and detailed summaries yet with more semantic mistakes that make them harder to understand (hence lower fluency scores). However, summaries by both systems are similarly preferred in terms of coherence. Also, we note that \textsc{PASS} utilizes a separately trained classifier on human-annotated summaries \citep{fabbri2021summeval} to rank candidate summaries, while our approach does not.

\paragraph{Input Content Fidelity.}
As was shown in \citet{falke2019ranking, tay2019red}, the ROUGE metric can be insensitive to hallucinations~\citep{maynez2020faithfulness}. However, hallucinations can lead to user aversion, and their reduction remains an open problem in summarization. To assess the input fidelity of generated summaries, we performed a human evaluation. Specifically, we used summaries produced by the adapter models (\adasumloo{5} and \adaqsumloo{5}), \textsc{FewSum}, \textsc{PASS}, and human-written (\textsc{Gold}). In each task (HIT), we presented both reviews and all summary sentences. We asked three workers to assess how well the content in summary sentences is supported by the reviews. The three following options were available. \textit{Full
  support}: all the content is reflected in the reviews;
\textit{Partial support}: only some content is reflected in the
reviews; \textit{No support}: content is not reflected in the reviews. The results, normalized by sentences, are shown in Table~\ref{table:content_support}.

\begin{table}[t]
\centering
\small
\begin{tabular}
{l c c c} \thickhline
& Full$\uparrow$ & Partial$\uparrow$  & No$\downarrow$ \\ \thickhline
\textsc{FewSum} & 47.56 & 24.39 & 28.05 \\
\textsc{PASS} & 60.70 & 31.84 & 7.46 \\ 
\adasumloo{5} & \textbf{78.97} & \textbf{15.48} & \textbf{5.56} \\ 
\adaqsumloo{5} & 72.69 & 20.37 & 6.94 \\ 
\thickhline
\end{tabular}
\caption{Input fidelity on the Amazon test set, normalized by sentences.}
\label{table:content_support}
\TablePad{}
\end{table}

First, we observe that \textsc{FewSum} hallucinates the most, potentially because it was not initialized with a pre-trained language model. Second, \textsc{PASS} improves input fidelity over \textsc{FewSum} yet substantially underperforms our adapter-based models. We also notice a slight decrease in input fidelity when the query is used. This is likely caused by more abstractive summaries generated by \adaqsumnoperc{}, we discuss it in Sec.~\ref{sec:abstractivness}.

%% file: content/sections/analysis.tex
\section{Analysis}
\label{sec:analysis}
\subsection{Query-based Pre-training}
\label{sec:query_based_summarization_results}

Query-based summarizers should generate summaries reflecting all aspects in user queries to be useful in practice. We investigated how summarizers learn this task in the few-shot regime with and without pre-training. We created test-time queries from gold summaries (indicated by $^{*}$) and input reviews. Further, we calculated the aspect recall (\textsc{AR}) score by counting aspect keywords in queries present in generated summaries. The results are shown in Table~\ref{table:auto_res_query_summ}. 

As indicated by low \textsc{AR} scores, without pre-training, the models miss many aspects in queries. The increase to nearly 100\% in AR suggests that pre-training is crucial for the task. The same trend remains when aspect keywords from reviews are used in queries.

\begin{table}[t!]
    \centering
    \small
    \begin{tabular}{l c c c c} \thickhline
    & R1 & R2 & RL & AR \\
    \thickhline
    \textsc{Full (100\%) + q}$^{*}$ & 40.52 & 10.96 & 25.06 & 59.84 \\
    \textsc{Full (100\%) + l1o + q}$^{*}$ & 42.65 & 11.53 & 26.82 & 96.39 \\ \hline
    \adaqsum{5}$^{*}$ & 41.04 & 11.08 & 25.46 & 60.64 \\
    \adaqsumloo{5}$^{*}$ & 43.84 & 13.41 & 27.31 & 97.19 \\ \hline
    \adaqsum{5} & 38.58 & 10.10 & 24.19 & 69.14 \\
    \adaqsumloo{5} & 38.53 & 10.52 & 26.06 & 98.78 \\ \hline

    \thickhline
    \end{tabular}
    \caption{Amazon test set ROUGE F1 for query-based summarization. Here, $^{*}$ indicates that queries were created from gold summaries; \textsc{AR} stands for aspect recall.}
    \label{table:auto_res_query_summ}
\end{table}

\subsection{Catastrophic Forgetting}
\label{sec:catastrophic_forgetting}
ROUGE scores in Table~\ref{table:auto_res_generic_summ} suggest that \loo{} pre-training is beneficial for the end task. However, fine-tuning on summaries can lead to the catastrophic forgetting of the acquired in-domain specifics from reviews. Because adapters have fewer parameters to optimize, we hypothesized that they might be more robust to this phenomenon. 

To test the hypothesis, we evaluated two models on the pre-training \loo{} pairs where a review is used as a summary, before and after fine-tuning on human-written summaries. For the first model, we optimized only adapters (5\%), both in pre-training and fine-tuning. And in the second case, we optimized the entire model. We used PPL to measure the model's confusion about the pre-training pseudo summaries, as shown in Table~\ref{table:catastrophic_forgetting}.  

The results demonstrate that the adapter-based model better preserves information about reviews after they are fine-tuned on summaries, as indicated by lower PPL scores. Our findings are also supported by \citet{yu2021AdaptSum}.

\begin{table}[t!]
\centering
\small
\begin{tabular}
{ l  l l } \thickhline
& PPL$\downarrow$ \\ \thickhline
\textsc{Full (100\%) + l1o} & 21.51 \\ 
\textsc{Full (100\%) + l1o + ft} & 34.87 (+13.36) \\ \hline
\adasumloo{5} & 19.69 \\ 
\adasumloo{5} \textsc{+ ft} & \textbf{28.45} (\textbf{+8.76}) & \\ 
\thickhline
\end{tabular}
\caption{Catastrophic forgetting evaluation on the Amazon pre-training task's validation set, before and after fine-tuning (\textsc{FT}).}
\label{table:catastrophic_forgetting}
\TablePad{}
\end{table}

\subsection{Abstractiveness}
\label{sec:abstractivness}

Abstracting information in reviews is important for practical applications~\citep{carenini-cheung-2008-extractive}. To investigate how well the models abstract, we computed the number of novel n-grams in generated summaries with respect to input reviews on the Amazon test set. The results in percentages are shown in Table~\ref{table:abstractivness}.

First, we observe that \textsc{FewSum} tends to produce the most abstractive summaries, followed by \textsc{PASS}. Second, \adaqsumnoperc{} has higher abstractivness than \adasumnoperc{}. We also observe that abstractiveness is inversely proportional to input faithfulness in Table~\ref{table:content_support}, in line with previous studies~\citep{durmus-etal-2020-feqa,dreyer-etal-2021-tradeoff}.

\begin{table}[t!]
    \centering
    \small
    \begin{tabular} {l c c c}
         \thickhline
          & 2-gram & 3-gram & 4-gram \\
         \thickhline
         \textsc{FewSum} & \textbf{78.63} & \textbf{95.59} & \textbf{98.74} \\
         \textsc{PASS} & 70.72 & 86.32 & 93.24 \\
         \adasumloo{5} & 55.47 & 78.24 & 86.78 \\
         \adaqsumloo{5} & 56.27 & 79.18 & 88.48 \\
         \thickhline
    \end{tabular}
    \caption{The abstractiveness of generated summaries in terms of novel n-grams on the Amazon test set.}
    \label{table:abstractivness}
    \TablePad{}
\end{table}

\subsection{Semantic Mistakes}
\label{sec:semantic_mistakes}

When a pre-trained model (with and without adapters) is fine-tuned on a handful of annotated samples, it often results in summaries with subtle semantic mistakes; see examples in Table~\ref{table:semantic_mistakes-1}. For instance, a 5\% adapter model generates a semantically contradicting fragment: \textit{`This \textbf{dead on arrival battery} is of good quality and holds a charge well.'}. 

We hypothesize that it is caused by the lack of in-domain knowledge, which we propose to learn via \loo{} (see Sec.~\ref{sec:self_sup_pretraining}). During a manual investigation, we observed that \loo{} pre-training substantially reduces semantic mistakes. This is also reflected in higher ROUGE scores in Table~\ref{table:auto_res_generic_summ}.

\input{content/examples/semantic_mistakes-1}

%% file: content/examples/semantic_mistakes-1.tex
\begin{table}[t!]
    \centering
 	\footnotesize 
    \begin{tabular}{ >{\arraybackslash} m{0cm} m{1.6cm} m{5.0cm}} 
 \thickhline 
 \multirow{2}{*}{} & \adasum{5} & \vspace{0.5em} This Thomas the Train costume is cute and functional. The size is perfect for a toddler or 2 year old and the candy pouch is large enough to carry a lot of candy. The costume can be customized with googly eyes, pumpkin patch, spiders, bats, or train tracks to make it more suitable for a child of any age. The fit is comfortable and the \textbf{fit quality is great}. The only complaint I have is \textbf{the quality of the felt fabric}. \vspace{0.5em} \\
 & \adasumloo{5} & \vspace{0.5em} This Thomas the Train costume is very cute and the size is perfect for a 2-year-old. The hat is thin and flimsy and the face is not 3D sculpted. The candy pouch is a nice feature and it looks like it will grow with the child. Overall, it's a cute costume and will be used for Halloween next year. \vspace{0.5em}\\ \hline
   \multirow{2}{*}{} & \adasum{5} & \vspace{0.5em} \textbf{This dead on arrival battery is of good quality and holds a charge well}. It is easy to install and is a great value for the money. \textbf{However, it may not hold a charge as advertised due to the plastic case bulging}. Overall, this product is highly recommended. \vspace{0.5em}\\
 & \adasumloo{5} & \vspace{0.5em} This battery is a great value for the price and works great. It is a good quality battery that can be used to replace a dead battery in an alarm system. The price is great and the quality of the product is good. The shipping was fast and the customer service was excellent. \\  \thickhline 
    \end{tabular}
    \caption{Adapter-based models (5\%) and their generated outputs with and without \loo{} pre-training. Semantic mistakes and disfluencies are highlighted in \textbf{bold}. }
    \label{table:semantic_mistakes-1}
    \TablePad{}
\end{table}

%% file: content/sections/related_work.tex
\section{Related Work}
\label{sec:related_work}

Extractive opinion summarization has been an active
area of research \citep{hu2004mining, ganesan2010opinosis, medhat2014sentiment, isonuma2019unsupervised, angelidis2020extractive}. For example, a more recent extractive method of \citet{angelidis2018summarizing} decouples the summarization procedure into multiple steps with separate models. Other earlier approaches~\citep{gerani2014abstractive, di2014hybrid} relied on text planners and templates, which, however, restrict the output. 

Abstractive opinion summarization is an emerging branch~\citep{chu2019meansum, amplayo2020unsupervised, brazinskas2020-unsupervised, bravzinskas2021learning}. Customer reviews were used to train unsupervised summarizers in \citet{amplayo2020unsupervised, brazinskas2020-unsupervised, isonuma2021unsupervised}. 
The few-shot model \textsc{FewSum}~\citep{brazinskas2020few} was also pre-trained on customer reviews before fine-tuning. In this work instead, we focus on pre-training adapters to avoid catastrophic forgetting and reduce computational and memory overheads. \textsc{OpinionDigest}~\citep{suhara-etal-2020-opiniondigest} proposes to aggregate opinions in a pipeline framework. We approach the problem end-to-end and rely on aspect keywords (e.g., price) instead of opinion phrases (e.g., good location). Controllability using input fragments (e.g., entities) and meta information (e.g., coarse-grained aspects) has received recent attention in various NLP domains~\citep{frermann-klementiev-2019-inducing, liu-chen-2021-controllable, narayan2021planning, elsahar-etal-2021-self}. In contrast to a related work, \textsc{SelfSum}~\citep{elsahar-etal-2021-self}, we use aspect keywords instead of generic tokens and consider a few-shot setup instead of unsupervised and a different model architecture. Also, planning was tackled in opinion summarizeration in \citet{amplayo2021unsupervised}. However, their approach is substantially less flexible, as the summary plan consists of an aspect and sentiment classes only. Query-based settings have received recent attention in the news domain~\citep{xu2020coarse, xu2021text}. Compared to a concurrent work on opinion summarization \textsc{AceSum}~\citep{amplayo2021aspect}, our approach does not require a trained aspect induction model, is few-shot instead of self-supervised, and benefits from a large collection of automatically created fine-grained aspects (a couple of thousands) instead of human annotated coarse-grained aspects (up to 18). Concurrently with our work,~\citet{poth2021pre} support our findings on the benefits of pre-training adapters for other tasks. 

%% file: content/sections/conclusion.tex
\section{Conclusions}
\label{sec:conclusion}

In this work, we improve few-shot learning for opinion summarization with adapters pre-training on customer reviews in the end task-specific manner. In this way, the model learns in-domain specifics, which reduces semantic mistakes in generated summaries. We show that our approach leads to more than 2.0 and 1.3 ROUGE-L points improvement over the entire  model's fine-tuning on the Amazon and Yelp datasets, respectively. Further, we propose a simple method for few-shot query-based summarization. The queries consist of aspect keywords reflecting potential user interests. We create these queries automatically and show that pre-training is crucial for the end task, significantly improving performance. Finally, in human evaluation, we demonstrate that the query-based model generates more coherent and less redundant summaries.

%% file: content/sections/acknowledgment.tex
\section*{Acknowledgments}
We would like to thank Jonathan Pilault, Leonardo Ribeiro, Sandeep Atluri, the Amazon AI teams for their useful feedback and discussions. Also, we would like to thank the anonymous reviewers for their insightful comments and suggestions. 

%% file: content/sections/ethics.tex
\section{Ethics Statement}
\label{sec:ethics}

We used only publicly available datasets. For human evaluation, we used a publicly available service (Amazon Mechanical Turk) to hire voluntary participants, requesting native speakers of English. The participants were fairly compensated, above the minimum hourly wage in their self-identified countries of residence.

%% file: content/sections/limitations.tex
\section{Limitations}

In this work, we explicitly focus on multi-document abstractive opinion summarization. However, our pre-training self-supervised method and fine-tuning techniques can be applied to a broader set of multi-document summarization domains (e.g., news) and can be considered in the future work. Also, while we tested our approach only with BART, we believe that it would work with other pre-trained encoder-decoder models, like PEGASUS~\citep{zhang2019pegasus}.

%% file: content/appendix/main.tex
\section{Appendices}
\label{sec:appendix}

\input{content/appendix/human_evaluation}

\input{content/appendix/best_worst}

\input{content/appendix/examples}

%% file: content/appendix/human_evaluation.tex
\subsection{Human Evaluation Setup}
\label{sec:he_setup}

To performed the human evaluation experiments described in Sec.~\ref{sec:human_eval}, we hired workers with 98\% approval rate, 1000+ HITS, Location: USA and the maximum score on a qualification test that we had designed. The test asked if the workers were native English speakers, and verified that they correctly understood the instructions of both the best-worst scaling and content support tasks. We paid the workers an approximate amount of \$12 per hour.

%% file: content/appendix/best_worst.tex
\subsection{Best-Worst Scaling Details}
\label{app:best_worst}

We performed human evaluation based on the Amazon test set using the AMT platform. We assigned 3 workers to each tuple containing summaries from different systems. We showed summaries and asked to select the best and worst one based on the criterion presented below.

\begin{enumerate}
    \item \textit{Fluency}: the summary sentences should be grammatically correct, easy to read and understand; 
    \item \textit{Coherence}: the summary should be well structured and well organized;
    \item \textit{Non-redundancy}: there should be no unnecessary repetition in the summary.
\end{enumerate}

%% file: content/appendix/examples.tex
\input{content/examples/ama_example1}

\input{content/examples/ama_example2}

%% file: content/examples/ama_example1.tex
\begin{table*}
    \centering
 	\footnotesize 
    \begin{tabular}{  >{\centering\arraybackslash} m{2.5cm} m{12cm}}
 \thickhline
    \textsc{Gold} & \vspace{0.5em} These transition tights are perfect for children sensitive to the tight sensation other tights have around the foot.  The material is soft and durable; they stand up well to both the rough nature of children, and the washing machine.  This product does tend to run slightly small, so purchasing one size up is recommended.
 \vspace{0.5em} \\ \hline
    \textsc{FewSum} & \vspace{0.5em} These tights are a great value for the price. The fit is true to size and the quality of the tights is very good. They are well made and will last a long time. They do run a little on the small side, so order a size up.
 \vspace{0.5em} \\ \hline
    \textsc{PASS} & \vspace{0.5em} These soft, breathable tights are great for transitioning from tap to ballet. They fit snugly around the body and stay in place when worn with ballet shoes. They are well made and well made, and can last longer than other tights available. The colors are beautiful and will definitely be purchasing again.
 \vspace{0.5em} \\ \hline
     \adasumloo{5} & \vspace{0.5em} These tights are soft and comfortable and fit well. They are durable and will last a long time. They can be worn with sandals or flip-flops. They do run small and should be ordered one size up to avoid squishing toes. The color is beautiful and the material is soft and durable.
 \vspace{0.5em} \\ \hline
    \adaqsumloo{5} & \vspace{0.5em} These tights are soft, comfortable, and durable. The \textbf{color} is beautiful and the \textbf{fit} is perfect for \textbf{tap} and ballet. They \textbf{fit} well and are durable enough to \textbf{wear} with flip-flops to class. They are recommended to order one \text{size} up if your child is chubby or slim.
 \vspace{0.5em} \\
     \thickhline
    \textsc{Review 1} & \vspace{0.5em} These are the perfect tights for my 5-year old. The tights are very well made and have already lasted several washings (hang dry). The color is beautiful, and my daughter loves that she can wear flip-flops to class like the big girls do. \vspace{0.5em} \\ \hline
    \textsc{Review 2} & \vspace{0.5em} my 3 year old fit into these perfectly. I love these tights, they are great for wearing sandals to dance class and then pulling them over her toes to put ballet slippers on. They are nice and soft and the pink color is pretty. Will purchase again. \vspace{0.5em} \\ \hline
    \textsc{Review 3} & \vspace{0.5em} These are my daughters preferred ballet tights. They fit well and don't squish her toes as much as some others. The convertible option is nice as she can wear flip flops to the studio with her tights. I like that they appear to be fairly durable. \vspace{0.5em} \\ \hline
    \textsc{Review 4} & \vspace{0.5em} Bought this for my little one to use for her ballet class. She's almost 4 and this fits perfectly. Transition tights give her the ability to pull up the foot area to around the ankles so that they don't get dirty when not wearing her shoes, but fit well and stay in place when pulled over her feet and used with ballet shoes on. \vspace{0.5em} \\ \hline
    \textsc{Review 5} & \vspace{0.5em} Great soft fabric, runs small a though. U should consider getting one size up to avoid having your daughter, or son if he's into ballet, have little circulation. Don't get me wrong, great product and material. \vspace{0.5em} \\ \hline
    \textsc{Review 6} & \vspace{0.5em} I purchased these tights for my 4 year old because she has a tap class immediately followed by an acro class. Tights fit well and were easy to transition to bare feet after tap. I can't comment on how they hold up, as we only had one week of dance so far. \vspace{0.5em} \\ \hline
    \textsc{Review 7} & \vspace{0.5em} This is my go to tight when my daughter needs new ones. I won't buy anything else. They are super comfy and last forever! She is 7 but very slim and these will fit her for a while but the 8-12 is a better fit for longevity purposes but they only had this size left and I needed them right away. \vspace{0.5em} \\ \hline
    \textsc{Review 8} & \vspace{0.5em} These are great tights. If your girl is chubby on the waist i think you should order a number up. they are a strong tihgts. My daugther has it for more than 4 months and hasn't broken them. She only complains some days about the waist, just when she has had a lot of milk or food recently. she's 3 years old. \vspace{0.5em} \\ \thickhline
    \end{tabular}
    \caption{Example summaries produced by different systems for an Amazon product. We highlight in \textbf{bold} keywords that were part of the query for \adaqsumloo{5}.}
\end{table*}

%% file: content/examples/ama_example2.tex
\begin{table*}
    \centering
 	\footnotesize 
    \begin{tabular}{  >{\centering\arraybackslash} m{2.5cm} m{12cm}}
 \thickhline
    \textsc{Gold} & \vspace{0.5em} The results with this scanner are sporadic at best- there are many problems with the software drivers and the stringent system requirements that come along with it. It does not work with the most recent versions of Windows, and the company charges for tech support calls.
 \vspace{0.5em} \\ \hline
    \textsc{FewSum} & \vspace{0.5em} This is a great scanner. It is easy to use and easy to set up. The software that comes with it does not work with Windows Vista. It would be nice if it was a little more user friendly, but for the price it can't be beat.
 \vspace{0.5em} \\ \hline
    \textsc{PASS} & \vspace{0.5em} The VuPoint Digital Scanner is very user friendly and easy to use. It can scan over 1200 35 mm negatives and about a dozen slides. It takes time and money to get the scanner to fit in the holders that come with it. The company also charges \$10.00 per phone call for tech support.
 \vspace{0.5em} \\ \hline
     \adasumloo{5} & \vspace{0.5em} The VuPoint Digital Scanner is easy to use and does a great job converting negatives to digital format. However, the software is not compatible with newer versions of Windows. The company ArcSoft charges \$10 per phone call for tech support. Overall, this product is not recommended.
 \vspace{0.5em} \\ \hline
    \adaqsumloo{5} & \vspace{0.5em} The VuPoint Digital Scanner does a great job of converting \textbf{photo} negatives to digital format. The \textbf{software} is easy to use and easy to install. However, the \textbf{image} bleaches out with too much light. A \textbf{replacement} unit is required. Overall, this product is recommended.
 \vspace{0.5em} \\
     \thickhline
    \textsc{Review 1} & \vspace{0.5em} I recently bought this film and slide scanner to scan my grandfather's slide collection. It bleaches out the image with too much light. I tried changing the settings to improve the image quality, but had no luck. The company ArcSoft charges \$10.00 per phone call for tech support. You are better off making the investment on a nicer quality scanner. \vspace{0.5em} \\ \hline
    \textsc{Review 2} & \vspace{0.5em} * * Not Reccommended * * Purchased as a gift in August. Opened a week ago. Spent the last week trying to get Win Xp to recognise the Vu Point scanner. Many drivers and reloads later all I have is a little black box with a red light and a message from windows that says 'USB Device Not Recognised'. \vspace{0.5em} \\ \hline
    \textsc{Review 3} & \vspace{0.5em} I used the VuPoint Digital Scanner to scan over 1,200 35 mm negatives and about a dozen slides and found this gadget a most user-friendly and efficient tool. I even managed to upload a few black and white negatives from 1963. I recommend the product highly. \vspace{0.5em} \\ \hline
    \textsc{Review 4} & \vspace{0.5em} While the software was good for Windows XP and Vista, I now have Windows 7 and would like to have software for the newer operating system. The company prefers to sell other products rather than update their software. I can't see recommending this product in today's market. \vspace{0.5em} \\ \hline
    \textsc{Review 5} & \vspace{0.5em} While most equipment will work with more modern versions of Windows than were available when manufactured this is not true with this scanner. Requires Windows XP means it won't work with earlier OR LATER. Its on its way back for a refund. \vspace{0.5em} \\ \hline
    \textsc{Review 6} & \vspace{0.5em} I found the VuPoint scanner not acceptable and I am still waiting for a replacement. My contacts with VuPoint were helpful but the equipment still did not produce acceptable images. My contact with the seller has been sporadic, at best, and a replacement unit has not been delivered.I an NOT anxious to deal with these providers again. \vspace{0.5em} \\ \hline
    \textsc{Review 7} & \vspace{0.5em} Product is very easy to use. Does a great job converting my slide and photo negatives to digital format. Touch-up and enhance program gave me just what I needed to clean up and enhance some of the scans, Company was great to work with!! \vspace{0.5em} \\ \hline
    \textsc{Review 8} & \vspace{0.5em} Its not worth the time it takes to get the negative to fit in the holders they give you. I'd much rather buy a hp flat bed scanner that lets you see the final photo image and not just an image of the negative. It takes to much time and isn't worth the money. \vspace{0.5em} \\ \thickhline
    \end{tabular}
    \caption{Example summaries produced by different systems for an Amazon product. We highlight in \textbf{bold} keywords that were part of the query for \adaqsumloo{5}.}
\end{table*}